\documentclass[final]{cvpr}

\makeatletter
\@namedef{ver@everyshi.sty}{}
\makeatother

\usepackage{times}
\usepackage{epsfig}
\usepackage{graphicx}
\usepackage{amsmath}
\usepackage{amssymb}

\usepackage{booktabs}       
\usepackage{amsfonts}       
\usepackage{nicefrac}       
\usepackage{microtype}      
\usepackage{subfig}
\usepackage{graphicx}
\usepackage{multirow}
\usepackage{xspace}
\usepackage{commath}
\usepackage{bm}
\usepackage{tikz}  

\usepackage{pgfplots}

\usepackage{soul}
\usepackage{colortbl}
\usepackage{xcolor}

\usepackage{footnote}
\usepackage{footmisc}
\usepackage{tablefootnote}

\pgfplotsset{compat=1.14}

\usepackage[pagebackref=true,breaklinks=true,colorlinks,bookmarks=false]{hyperref}

\def\cvprPaperID{5926} 
\def\confYear{CVPR 2021}

\begin{document}

\title{StickyPillars: Robust and Efficient Feature Matching on Point Clouds\\
       using Graph Neural Networks}

\author{Kai Fischer$^1$, Martin Simon$^1$, Florian Ölsner$^2$, Stefan Milz$^{2,4}$, Horst-Michael Groß$^3$, Patrick Mäder$^4$\vspace*{8pt}\\
$^1$ Valeo Schalter und Sensoren GmbH, Kronach, Germany\\
$^2$ Spleenlab GmbH, Saalburg-Ebersdorf, Germany\\
$^3$ Neuroinformatics and Cognitive Robotics Lab, Ilmenau University of Technology, Germany\\
$^4$ Software Engineering for Safety-Critical Systems, Ilmenau University of Technology, Germany\\
{\tt\small $\{$kai.fischer, martin.simon$\}$@valeo.com}
}

\maketitle

\newcommand{\lidar}{LiDAR\xspace}
\newcommand{\lidars}{LiDARs\xspace}
\newcommand{\trainone}{\mathcal{T}_1}
\newcommand{\traintwo}{\mathcal{T}_5}
\newcommand{\trainthree}{\mathcal{T}_{10}}
\newcommand{\validone}{\mathcal{V}_1}
\newcommand{\validtwo}{\mathcal{V}_5}
\newcommand{\validthree}{\mathcal{V}_{10}}
\newcommand{\GNN}{\textit{Graph Neural Network}\xspace}
\newcommand{\GNNs}{\textit{Graph Neural Networks}\xspace}
\newcommand{\mcP}{\mathcal{P}}
\newcommand{\pcO}{{\bar{\mathcal{P}}}}
\newcommand{\uK}{{\mathrm{K}}}
\newcommand{\uL}{{\mathrm{L}}}
\newcommand{\pcK}{\mathcal{P}^{\mathrm{K}}}
\newcommand{\pcL}{\mathcal{P}^{\mathrm{L}}}
\newcommand{\kpK}[1]{\bm{\pi}^{\mathrm{K}}_{#1}}
\newcommand{\kpL}[1]{\bm{\pi}^{\mathrm{L}}_{#1}}
\newcommand{\hkpK}[1]{\hat{\bm{\pi}}^{\mathrm{K}}_{#1}}
\newcommand{\hkpL}[1]{\hat{\bm{\pi}}^{\mathrm{L}}_{#1}}
\newcommand{\pK}[1]{\bm{x}^{\mathrm{K}}_{#1}}
\newcommand{\pL}[1]{\bm{x}^{\mathrm{L}}_{#1}}

\newcounter{savefootnote}
\newcounter{symfootnote}
\newcommand{\symfootnote}[1]{%
   \setcounter{savefootnote}{\value{footnote}}%
   \setcounter{footnote}{\value{symfootnote}}%
   \ifnum\value{footnote}>8\setcounter{footnote}{0}\fi%
   \let\oldthefootnote=\thefootnote%
   \renewcommand{\thefootnote}{\fnsymbol{footnote}}%
   \footnote{#1}%
   \let\thefootnote=\oldthefootnote%
   \setcounter{symfootnote}{\value{footnote}}%
   \setcounter{footnote}{\value{savefootnote}}%
}

\begin{abstract}
Robust point cloud registration in real-time is an important prerequisite for many mapping and localization algorithms. Traditional methods like ICP tend to fail without good initialization, insufficient overlap or in the presence of dynamic objects. Modern deep learning based registration approaches present much better results, but suffer from a heavy run-time. We overcome these drawbacks by introducing StickyPillars, a fast, accurate and extremely robust deep middle-end 3D feature matching method on point clouds. It uses graph neural networks and performs context aggregation on sparse 3D key-points with the aid of transformer based multi-head self and cross-attention. The network output is used as the cost for an optimal transport problem whose solution yields the final matching probabilities. The system does not rely on hand crafted feature descriptors or heuristic matching strategies.
We present state-of-art art accuracy results on the registration problem demonstrated on the KITTI dataset while being four times faster then leading deep methods. Furthermore, we integrate our matching system into a LiDAR odometry pipeline yielding most accurate results on the KITTI odometry dataset. Finally, we demonstrate robustness on KITTI odometry. Our method remains stable in accuracy where state-of-the-art procedures fail on frame drops and higher speeds.
\end{abstract}

\section{Introduction}

Point cloud registration, the process of finding a spatial transformation aligning two point clouds, is an essential computer vision problem and a precondition for a wide range of tasks in the domain of real-time scene understanding or applied robotics, such as odometry, mapping, re-localization or SLAM. New generations of 3D sensors, like depth cameras or \lidars (light detection and ranging), as well as multi-sensor setups provide substantially more fine-grained and reliable data enabling dense range perception at a large field of view. These sensors substantially increase the expectations on point cloud registration and an exact matching of feature correspondences.

\begin{figure*}[!t]
\centering
\includegraphics[width=1\textwidth]{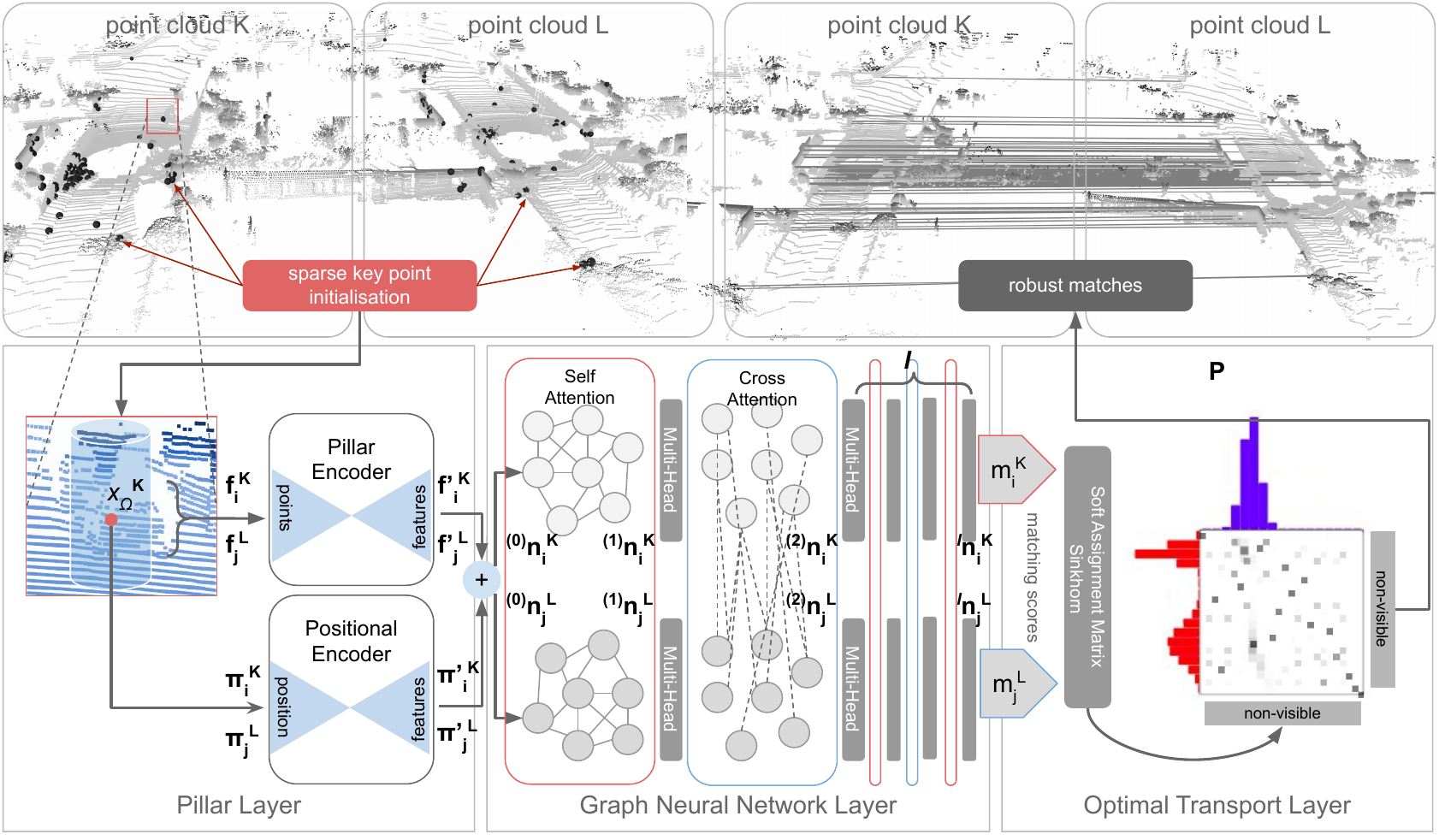}
\caption{\textbf{StickyPillars Architecture} is composed by three layers: 1. \textit{Pillar Layer}, 2. \textit{\GNN layer} and 3. \textit{Optimal Transport layer}. With the aid of 1, we learn 3D features (\textit{pillar encoder}) and spatial clues (\textit{positional encoder}) directly. In 2 \textit{Self-} and \textit{Cross} Multi-Head Attention is performed in a graph architecture for contextual aggregation. The resulting matching scores are used in 3 to generate an assignment matrix for key-point correspondences via numerical optimal transport. }
\label{fig:architecture}
\end{figure*}

State-of-the-art 3D point cloud registration employs locally describable features in a global optimization step \cite{legoloam2018,Zhang-2014-7903,lin2019loam_livox}. Most methods do not rely on modern machine learning algorithms, although they are part of the best performing approaches on odometry challenges like KITTI \cite{geiger2012we}. In contrast, recent research for point cloud processing, e.g. classification and segmentation \cite{qi2017pointnet, qi2017pointnet++, lang2019pointpillars, zhou2018voxelnet}, relies on neural networks and promises substantial improvements for registration, mapping and odometry \cite{engel2019deeplocalization, li2019net}. 
The limitation of all none neural network-based odometry and mapping methods is that they perform odometry estimation using a global rigid body operation. Those approaches assume many static objects within the environment and proper viewpoints. However, real world measurements are generally unstable under challenging situations, e.g. many dynamic objects or widely varying viewpoints and small overlapping areas. Hence, the mapping quality itself is suffering from artifacts (blurring) and is often not evaluated qualitatively. To overcome these limitations, we propose StickyPillars a novel registration approach for point clouds utilizing graph neural networks. Inspired by \cite{detone2018superpoint, vaswani2017attention}, our approach computes feature correspondence rather than end-to-end odometry estimations. We demonstrate StickyPillars's robust real-time registration capabilities (see Fig.~\ref{fig:architecture}) and its confidence under challenging conditions, such as dynamic environments, challenging viewpoints and small overlapping areas. We evaluate our technique on the KITTI odometry benchmark \cite{geiger2012we} and significantly outperform state-of-the-art frame to frame matching approaches e.g. the one used in LOAM\cite{Zhang-2014-7903}. Those improvements enable more precise odometry estimation for applied robotics. 

\section{Related Work}

\textbf{Deep Learning on point clouds} is a novel field of research \cite{chen2017multi, simon2018complex, simon2019complexer}. The fact that points are typically stored in unordered sets and the need for viewpoint invariance prohibits the direct use of classical CNN architectures. Existing solutions tackle this problem by converting the point cloud into a voxel grid \cite{zhou2018voxelnet}, by projecting it to a sphere \cite{milioto2019iros, li2019net} or by directly operating on the set using well designed network architectures \cite{qi2017pointnet, qi2017pointnet++}. 

\textbf{Point cloud registration} aims to find the relative rigid 3D transformation that aligns two sets of points representing the same 3D structure. Most established methods try to find feature correspondences in both point clouds. The transformation is found by minimizing a distance metric using standard numerical solvers. A simple but powerful approach called iterative closest point (ICP) was exhaustively investigated by \cite{10.1109/34.121791, Zhang1992icp, Rusinkiewicz2001EVO} and adapted in a wide range of applications \cite{10.1007/978-3-540-24672-5_44,legoloam2018,Zhang-2014-7903,lin2019loam_livox}. Here each point is a feature which is matched to its closest neighbor in the other point cloud in an iterative process. ICP's convergence and runtime highly depends on the initial relative pose, the matching accuracy and the overlap \cite{Rusinkiewicz2001EVO}. This problem can be avoided by using global feature matching approaches \cite{rusu2009fast, Lu_2019_ICCV, wang2019deep, Aoki_2019_CVPR, Li_2019_ICCV} in combination with solvers like RANSAC \cite{fischler1981random, raguram2008comparative} that are robust to outlier correspondences. Alternatively the transformation can be directly estimated using a neural network as proposed by \cite{li2019net}. Scene flow based methods pursue a different approach by estimating point-wise translational motion vectors instead of a single rigid transformation \cite{dewan2016rigid, ushani2017learning, liu2019flownet3d, qi2017pointnet++, gu2019hplflownet, liu2019meteornet}. These methods naturally handle dynamic and non-rigid movements but are computationally quite demanding.

\textbf{Lidar odometry estimation} is a typical task that involves point cloud registration in real-time. Many of the aforementioned methods \cite{Aoki_2019_CVPR, Li_2019_ICCV, gojcic2020learning, bai2020d3feat, Lu_2019_ICCV}, especially those using deep neural networks, are not fast enough to deal e.g. with lidar sensors that run at 10-15 Hz. LOAM \cite{Zhang-2014-7903}, the currently best performing approach on the KITTI odometry benchmark \cite{geiger2012we} uses a variant of ICP that operates on a sparse set of feature points located on edge and surface patches. The feature extraction exploits the special scan structure of an rotating lidar sensor. After a rough pairwise matching the point clouds are accurately aligned to a local map. There exist several variants and improvements \cite{lin2019loam_livox, legoloam2018}. In the work by \cite{li2019net} it is shown that the pairwise matching step can be achieved by an appropriately sized CNN, but their method wasn't able to reduce the error any further.


\textbf{Feature based matching} is more widely used in the domain of image processing with prominent approaches, such as FLANN \cite{muja2009fast} and SIFT \cite{lowe2004distinctive}. The fundamental approach consists of several steps, point detection, feature calculation and matching. Such models based on neighborhood consensus were evaluated by \cite{bian2017gms, sattler2009scramsac, tuytelaars2000wide, cech2010efficient} or in a more robust way combined with a solver called RANSAC \cite{fischler1981random, raguram2008comparative}. Recently, deep learning based approaches, i.e. convolutional neural networks (CNNs), were proposed to learn local descriptors and sparse correspondences \cite{dusmanu2019d2, ono2018lf, revaud2019r2d2, yi2016lift}. However, these approaches operate on sets of matches and ignore the assignment structure. In contrast, \cite{sarlin2019superglue} focuses on bundling aggregation, matching and filtering based on novel \GNNs.

\section{The StickyPillars Architecture}
The idea behind StickyPillars is the development of a robust-point cloud registration and matching algorithm to replace standard methods (e.g. ICP) as most common matcher in applied robotics and computer vision algorithms like odometry, mapping or SLAM.
The 3D point cloud features (\textit{pillars}) are flexible and fully composed by learnable parameters. \cite{lang2019pointpillars} and \cite{zhou2018voxelnet} have proposed a 3D feature learning mechanism for perception tasks. We transform the concept to feature learning within a matching pipeline, but only using sparse sets of key points to ensure real-time capability and leanness. We propose an architecture using graph neural networks to learn geometrical context aggregation of two point sets in an end-to-end manner. The overall architecture is composed by three important layers: 1. \textit{Pillar Layer}, 2. \textit{\GNN layer} and 3. \textit{Optimal Transport layer} (see Fig.~\ref{fig:architecture}).

\textbf{Problem description} 
Let $\pcK$ and $\pcL$ be two point clouds to be registered. 
The key-points of those point clouds will be denoted as $\kpK{i}$ and $\kpL{j}$ with
$\{\kpK{0}, \ldots, \kpK{n}\} \subset \pcK$ and $\{\kpL{0}, \ldots, \kpL{m}\} \subset \pcL$, 
while other points will be defined as $\pK{k}\in\pcK$ and $\pL{l}\in\pcL$.
Each key-point with index $i$ is associated to a \textit{point pillar}, which can be pictured as a cylinder with an endless height, having a centroid position $\kpK{i}$ and a center of gravity $\hkpK{i}$. All points ($\mathcal{P}_i^{\mathrm{K}}$) within a pillar $i$ are associated with a pillar feature stack $\mathbf{f}_i^\uK \in \mathbb{R}^D$, with $D$ as pillar encoder input depth. The same applies for $\kpL{j}$. $\mathbf{c}_{i,j}$ and $\mathbf{f}_{i,j}$ compose the input for the graph.
The overall goal is to find partial assignments $\langle \kpK{i}, \kpL{j} \rangle$ for the optimal re-projection $\pcO$ with $\pcO:=f_{\kpL{j}\rightarrow \kpK{i}}(\pcL) \approx \pcK$. 

\subsection{Pillar Layer}
\label{section:pillar_layer}
\textbf{Key-Point Selection} is the initial part of the pillar layer with the aim to describe a dense point set with a sparse significant subset of key-points to ensure real-time capability.
Most common 3D sensors deliver dense point clouds having more than 120k points. Similar to \cite{Zhang-2014-7903}, we place the centroid pillar coordinates on sharp edges and planar surface patches as areas of interest. A smoothness term $c$ identifies smooth or sharp areas. For a point cloud $\pcK$ the smoothness term $c^\uK$ is defined by:

\begin{equation}
\label{formula_keypoint_selection}
\begin{aligned}
c^\uK = \frac{1}{\abs{\mathcal{S}} \cdot \norm{\pK{k}}} \cdot
\norm{\sum_{k' \in \mathcal{S}, k' \ne k} \left( \pK{k} - \pK{k'} \right) }
\end{aligned}
\end{equation}

where $k$ and $k'$ being point indices within the point cloud $\pcK$ having coordinates $\pK{k}, \pK{k'} \in \mathbb{R}^3$. $\mathcal{S}$ is a set of neighboring points of $k$ and $\abs{\mathcal{S}}$ is the cardinality of $\mathcal{S}$. With the aid of the sorted smoothness factors in $\pcK$, we define two thresholds $c_{\text{min}}^\uK$ and $c_{\text{max}}^\uK$ to pick a fixed number $n$ of key-points $\kpK{i}$ in sharp $c_k^\uK > c_{\text{max}}^\uK$ and planar regions $c_k^\uK < c_{\text{min}}^\uK$. This is also repeated for the target point-set with $c^\uL$ on $\pcL$ selecting $m$ points with index $j$.

\begin{figure*}[!t]
\centering
\includegraphics[width=\textwidth]{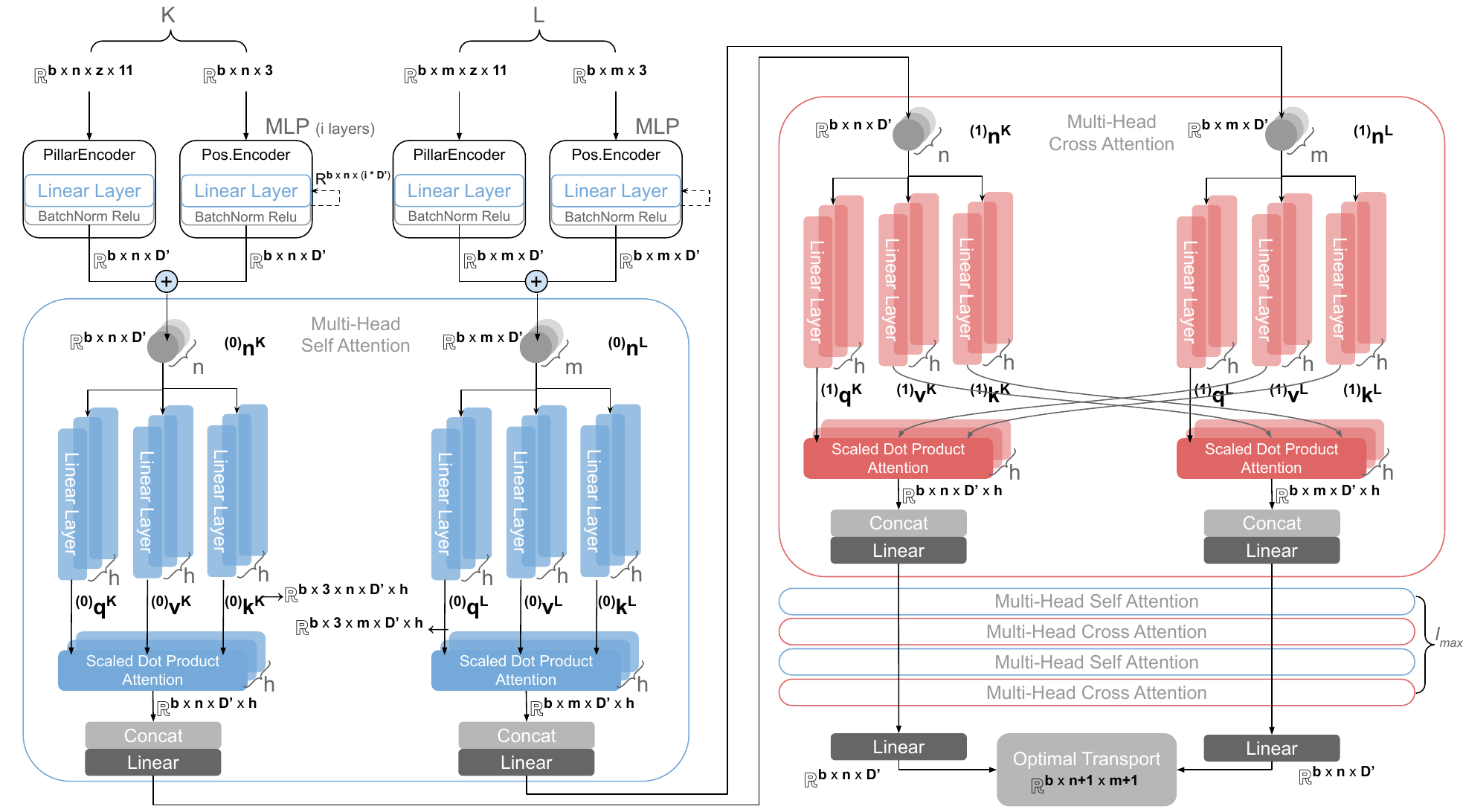}
\caption{\textbf{The StickyPillars Tensor Graph} identifies the data flow throughout the network architecture especially during \textit{self-} and \textit{cross} attention, where $b$ describes the batch-size, $n$ and $m$ the number of pillars, $h$ is the number of heads and $\mathrm{l}_{\mathrm{max}}$ the maximum layer depth. $D'$ is the feature depth per node. The result is an assignment matrix $\mathbf{P}$ with an extra column and row for invisible pillars.}
\label{fig:flow}
\end{figure*}

\textbf{Pillar Encoder} is designed to learn features in 3D inspired by \cite{qi2017pointnet, lang2019pointpillars}. Any selected key-point, $\kpK{i}$ and $\kpL{j}$, is associated with a \textit{point pillar} $i$ and $j$ describing a set of specific points $\mathcal{P}_i^\mathrm{K}$ and $\mathcal{P}_j^\mathrm{L}$. We sample points into a pillar using an 2D euclidean distance function (x,y plane) assuming a pillar alignment along the z coordinate (vertical direction) using a projection function $g \rightarrow [x,y,z] = [x,y]$:

\begin{equation}
\mathcal{P}_i^\mathrm{K} := \{\pK{0},\pK{\Omega}, ..., \pK{\mathrm{z}}\} \quad \norm{g(\kpK{i})-g(\pK{\Omega})} < \mathrm{d}
\end{equation}

Similar equations apply for $\mathcal{P}_j^\mathrm{L}$. Due to a fixed input size of the pillar encoder, we draw a maximum of $z$ points per pillar, where $z = 100$ is used in our experiments. $d$ is the distance threshold defining the pillar radius (e.g. 50\,cm). To enable efficient computation, we organized point clouds within a $k$-d tree \cite{Bentley_1975}. Based on $\kpK{i}$ the $z$ closest samples $\pK{\Omega}$ are drawn into the pillar $\mathcal{P}_i^\mathrm{K}$, whereas points with a projection distance greater $d$ were rejected. 

To compose a sufficient feature input stack for the \textit{pillar encoder} $\mathbf{f}_i^\uK \in \mathbb{R}^{D}$, we stack for each sampled point $\pK{\Omega}$ with $\Omega\in\set{1,\ldots,z}$ in the style of \cite{lang2019pointpillars}:
\begin{equation}
\begin{aligned}
\mathbf{f}_i^\uK&=\left\{
\Big\lbrack
\pK{\Omega}, i^{\mathrm{K}}_{\Omega}, (\pK{\Omega} - \hat{\bm{\pi}}^{\mathrm{K}}_{i}),  \|\pK{\Omega}\|_2, (\pK{\Omega} - \kpK{i})
\Big\rbrack,\ldots
\right\}
\end{aligned}
\end{equation}
$\pK{\Omega}\in \mathbb{R}^3$ denotes sample points' coordinates $(x,y,z)^{\mathrm{T}}$. $\bm{i}^{\mathrm{K}}_{\Omega} \in \mathbb{R}$ is a scalar and represents the intensity value (e.g. LiDAR reflectance), $(\pK{\Omega} - \hat{\bm{\pi}}^{\mathrm{K}}_{i}) \in \mathbb{R}^3$ being the difference to the pillar's center of gravity and $(\pK{\Omega} - \bm{\pi}^{\mathrm{K}}_{i}) \in \mathbb{R}^3 $ is the difference to the pillar's key-point. $\|\pK{\Omega}\|_2  \in \mathbb{R}$ is the L2 norm of the point itself. This leads to an overall input depth $D = z \times 11$.
The pillar encoder is a single linear projection layer with shared weights across all pillars and frames followed by a batchnorm and a ReLU layer with an output depth of $D'$ (e.g. 32 in our experiments) and $\mathbf{f'}_j^\uK, \mathbf{f'}_i^\uK  \in \mathbb{R}^{D'}$:
\begin{equation}
\begin{split}
\mathbf{f'}_i^\uK &= \mathrm{W}_\mathbf{f} \cdot \mathbf{f}_i^\uK \quad \forall i\in\set{1,\ldots,n} \\
\mathbf{f'}_j^\uL &= \mathrm{W}_\mathbf{f} \cdot \mathbf{f}_j^\uL \quad \forall j\in\set{1,\ldots,m}
\end{split}
\end{equation}

The aim of the Positional Encoder is learning geometrical aggregation using a context without applying pooling operations. The positional encoder is inspired by \cite{qi2017pointnet} and utilizes a single multi-layer-perceptron (MLP) shared across $\pcL$ and $\pcK$ such as all pillars including batchnorm and ReLU. From the centroid coordinates $\kpK{i}, \kpL{j} \in \mathbb{R}^3$, we calculate positional features via MLP with depth of $D'$ and $\bm{\pi'}_{i^\uK}, \bm{\pi'}_{j^\uL} \in \mathbb{R}^{D'}$:
\begin{equation}
\begin{split}
\bm{\pi'}_{i^\uK} &= \mathrm{MLP}_{\bm{\pi}}(\kpK{i}) \quad  \forall i\in\set{1,\ldots,n} \\
\bm{\pi'}_{j^\uL} &= \mathrm{MLP}_{\bm{\pi}}(\kpL{j}) \quad  \forall j\in\set{1,\ldots,m} 
\end{split}
\end{equation}

\subsection{Graph Neural Network Layer}
The \textbf{Graph Architecture} relies on two complete graphs $\mathcal{G}^\mathrm{L}$ and $\mathcal{G}^\mathrm{K}$, whose nodes are related and equivalent to the pillars quantity. The initial $^{(0)}\bm{n}_i^\mathrm{K}, ^{(0)}\bm{n}_j^\mathrm{L}$ node conditions are denoted as:

\begin{equation}
\begin{split}
^{(0)}\bm{n}_i^\mathrm{K} &= \mathbf{\pi'}_i^\uK + \mathbf{f'}_i^\uK \\
^{(0)}\bm{n}_j^\mathrm{L} &= \mathbf{\pi'}_j^\uL + \mathbf{f'}_j^\uL \qquad  ^{(0)}\bm{n}_i^\mathrm{K}, ^{(0)}\bm{n}_j^\mathrm{L} \in \mathbb{R}^{D'}
\end{split}
\end{equation}

The overall composed graph $(\mathcal{G}^\mathrm{L}, \mathcal{G}^\mathrm{K})$ is a \textit{multiplex} graph inspired by \cite{mucha2010community, nicosia2013growing}. It is composed by intra-frame edges, i.e. \textit{self} edges connecting each key-point within $\mathcal{G}^\mathrm{L}$ and each key-point within $\mathcal{G}^\mathrm{K}$ respectively. Additionally, to perform global matching using context aggregation inter-frame edges are introduced, i.e. \textit{cross} edges that connect all nodes of $\mathcal{G}^\mathrm{K}$ with $\mathcal{G}^\mathrm{L}$ and vice versa.

\textbf{Multi-Head Self- and Cross-Attention} allows us to integrate contextual cues intuitively and increase its distinctiveness considering its spatial and 3D relationship with other co-visible pillars, such as those that are salient, self-similar or statistically co-occurring \cite{sarlin2019superglue}. 
An attention function $\mathcal{A}$ \cite{vaswani2017attention} is a mapping function of a query and a set of key-point pairs to an output, with query $q$, keys $k$, and values $v$ being vectors. We define attention as: 
\begin{equation}
\begin{aligned}
\mathcal{A}(\bm{q},\bm{k},\bm{v})&=\mathrm{softmax}\left(\frac{\bm{q}^\mathrm{T}\cdot \bm{k}}{\sqrt{D'}}\right)\cdot \bm{v}
\end{aligned}
\end{equation}
where $D'$ describes the feature depth analogous to the depth of every node. We apply the multi-head attention function to each node $^{\mathrm{l}}\bm{n}_i^\mathrm{K}$, $^{\mathrm{l}}\bm{n}_j^\mathrm{L}$ at state $\mathrm{l}$ calculating its next condition $\mathrm{l}+1$. The node's conditions $\mathrm{l}\in\{0,\mathrm{l},...,\mathrm{l}_{\mathrm{max}}\}$ are represented as network layers to propagate information to the graph:

\begin{equation}
\begin{split}
{}^{(\mathrm{l}+1)}\bm{n}_i^\mathrm{K} &= {}^{(\mathrm{l})}\bm{n}_i^\mathrm{K} + {}^{(\mathrm{l})}\mathcal{M}^{\mathrm{K}}(\bm{q}_i^\mathrm{K},\bm{v}_\alpha^\mathrm{\Omega},\bm{k}_\alpha^\mathrm{\Omega}) \\
{}^{(\mathrm{l}+1)}\bm{n}_j^\mathrm{L} &= {}^{(\mathrm{l})}\bm{n}_j^\mathrm{L} + {}^{(\mathrm{l})}\mathcal{M}^{\mathrm{L}}(\bm{q}_j^\mathrm{L},\bm{v}_\beta^\mathrm{\Omega},\bm{k}_\beta^\mathrm{\Omega})
\end{split}
\end{equation}
We alternate the indices for $\alpha$ and $\beta$ to perform \textit{self} and \textit{cross} attention alternately with increasing depth of $\mathrm{l}$ through the network, where the following applies $\mathrm{\Omega} \in \{\mathrm{K}, \mathrm{L}\}$:
\begin{equation}
\begin{aligned}
\alpha, \beta&:= 
    \begin{cases}
      i, j &\text{if}~\mathrm{l} \equiv \text{even} \\
      j, i &\text{if}~\mathrm{l} \equiv \text{odd}
    \end{cases}
\end{aligned}
\end{equation}
The multi-head attention function is defined as:
\begin{equation}
\begin{aligned}
 {}^{(\mathrm{l})}\mathcal{M}^{\mathrm{K}}(\bm{q}_i^\mathrm{K},\bm{v}_\alpha^\mathrm{\Omega},\bm{k}_\alpha^\mathrm{\Omega}) =  {}^{(\mathrm{l})}\mathrm{W}_0 \cdot  {}^{(\mathrm{l})}(\mathrm{head}_1^\mathrm{K}\|...\|\mathrm{head}_h^\mathrm{K})
\end{aligned}
\end{equation}
with $\|$ being the concatenation operator. A single head is composed by the attention function:
\begin{equation}
\begin{split}
 {}^{(\mathrm{l})}\mathrm{head}_h^\mathrm{K} &= {}^{(\mathrm{l})}\mathcal{A}(\bm{q}_i^\mathrm{K},\bm{v}_\alpha^\mathrm{\Omega},\bm{k}_\alpha^\mathrm{\Omega}) \\
 &= {}^{(\mathrm{l})}\mathcal{A}(\mathrm{W}_{1h}\cdot\bm{n}_i^\mathrm{K},\mathrm{W}_{2h}\cdot\bm{n}_\alpha^\mathrm{\Omega},\mathrm{W}_{3h}\cdot\bm{n}_\alpha^\mathrm{\Omega})
\end{split}
\end{equation}
 The multi-head attention function is also defined for $ ^{(\mathrm{l})}\mathcal{M}^{\mathrm{L}}$. All weights ${}^{(\mathrm{l})}W_0,{}^{(\mathrm{l})}W_{11} .. {}^{(\mathrm{l})}W_{3h}$ are shared throughout all pillars and both graphs $(\mathcal{G}^\mathrm{L}, \mathcal{G}^\mathrm{K})$ within a single layer $\mathrm{l}$.

\textbf{Final predictions} are computed by the last layer within the Graph Neural Network and designed as single linear projection with shared weights across both graphs $(\mathcal{G}^\mathrm{L}, \mathcal{G}^\mathrm{K})$ and pillars:
\begin{equation}
\begin{split}
\mathbf{m}_i^\uK= \mathrm{W}_\mathbf{m} \cdot {}^{(\mathrm{l}_{\mathrm{max}})}\bm{n}_i^\mathrm{K} \\
\mathbf{m}_j^\uL= \mathrm{W}_\mathbf{m} \cdot {}^{(\mathrm{l}_{\mathrm{max}})}\bm{n}_j^\mathrm{L} \\ \mathbf{m}_j^\uL, \mathbf{m}_i^\uK \in \mathbb{R}^{D'}
\end{split}
\end{equation}

\subsection{Optimal Transport Layer}

Following the approach by \cite{sarlin2019superglue} the final matching is performed in two steps. First a score matrix $\mathbf{M} \in \mathbb{R}^{n\times m}$ is constructed by computing the unnormalized cosine similarity between each pair of features:
\begin{equation}
\begin{split}
\mathbf{M} &= (\mathbf{m}^\uK)^{\textrm{T}} \cdot \mathbf{m}^\uL,\\
\mathbf{m}^\uK &= [\mathbf{m}^\uK_1, \dots, \mathbf{m}^\uK_n], \; \mathbf{m}^\uL = [\mathbf{m}^\uL_1, \dots, \mathbf{m}^\uL_m]
\end{split}
\end{equation}
In the second step a soft-assignment matrix $\mathbf{P} \in \mathbb{R}^{(n + 1)\times (m + 1)}$ is computed that contains matching probabilities for each pair of features. Each row and column of $\mathbf{P}$ corresponds to a key-point in $\pcK$ and $\pcL$ respectively.
The last column and the last row represent an auxiliary \textit{dustbin} point to account for unmatched features.  Accordingly $\mathbf{M}$ is extended to a matrix $\mathbf{\bar{M}} \in \mathbb{R}^{(n + 1)\times (m + 1)}$ with all new elements initialized using a learnable parameter $W_v$. Finding the optimal assignment then corresponds to maximizing the sum $\sum_{i, j} \mathbf{\bar{M}}_{i, j} \mathbf{P}_{i, j}$ subject to the following constraints:
\begin{equation}
\begin{split}
    \sum_{i = 1}^{n + 1} \mathbf{P}_{i, j} = 
        \begin{cases}
            1 & \textrm{for } 1 \leq j \leq m \\
            n & \textrm{for } j = m + 1
        \end{cases} \\
    \sum_{j = 1}^{m + 1} \mathbf{P}_{i, j} = 
        \begin{cases}
            1 & \textrm{for } 1 \leq i \leq n \\
            m & \textrm{for } i = n + 1
        \end{cases}
\end{split}
\end{equation}
This represents an optimal transport problem \cite{sinkhorn1967concerning,vallender1974calculation,cuturi2013sinkhorn} which can be solved in a fast and differentiable way using a slightly modified version of the Sinkhorn algorithm \cite{sinkhorn1967concerning, cuturi2013sinkhorn}. Let $\mathbf{r}_i$ and $\mathbf{c}_j$ denote the $i^{\textrm{th}}$ row and $j^{\textrm{th}}$ column of $\mathbf{\bar{M}}$ respectively. A single iteration of the algorithm consists of 2 steps:
\begin{enumerate}
    \item $^{(t + 1)}\mathbf{r}_i \gets ~^{(t)}\mathbf{r}_i - \log \sum_j e^{\mathbf{r}_{i_j} - \alpha}$, with $\alpha = \log m$ for $i = n$ and $\alpha = 0$ otherwise
    \item $^{(t + 1)}\mathbf{c}_j \gets ~^{(t)}\mathbf{c}_j - \log \sum_i e^{\mathbf{c}_{j_i} - \beta}$, with $\beta = \log n$ for $j = m$ and $\beta = 0$ otherwise
\end{enumerate}
After $T = 100$ iterations we obtain $\mathbf{P} = \exp \left( ^{(T)}\mathbf{\bar{M}}\right)$. The overall tensor graph is shown in Fig~\ref{fig:flow} including architectural details from the pillar layer to the optimal transport layer.

\subsection{Loss}

The overall architecture with its three layer types: \textit{Pillar Layer, Graph Neural Network Layer} and \textit{Optimal Transport Layer} is fully differentiable. Hence, the network is trained in a supervised manner. The ground truth being the set $\mathcal{GT}$ including all index tuples $(i,j)$ with pillar correspondences in our datasets accompanied by unmatched labels $(\bar{n},j)$ and $(i, \bar{m})$, with $(\bar{n}, \bar{m})$ being redundant. We consider a negative log-likelihood loss $\mathcal{L}_\mathrm{NLL} = -\sum_{i,j\in \mathcal{GT}}\log\mathbf{P}_{ij}$.

\section{Experiments}
To present the performance of our method we separated the experiments into two subsections. First we validate the quality of StickyPillars in a point cloud registration task where we compare our results to geometry and DNN based state-of-the-art approaches. Subsequently we show how StickyPillars can be deployed as middle-end inside LiDAR odometry and mapping approaches by replacing the standard odometry estimation step to reduce drift and instabilities. We compare the performance on the KITTI Odometry benchmark to state-of-the-art methods like LOAM~\cite{zhang2017low} and LO-Net~\cite{li2019net}. Finally we demonstrate the robustness of StickyPillars simulating high speed scenarios by skipping certain amounts of frames on scenes from the KITTI odometry dataset.

\textbf{Model configuration:} For key-point extraction, we used variable $c_{\text{min}}$ and $c_{\text{max}}$ to achieve $n=m=500$ key-points $\bm{\pi}_i$ as inputs for the \textit{pillar layer}. Each \textit{point pillar} is sampled with up to $z=128$ points $\bm{x}_{\Omega}$ using an Euclidean distance threshold of $d=0.5\,\mathrm{m}$. Our implemented feature depth is $D'=32$. The key-point encoder has five layers with the dimensions set to $32,64,128,256$ channels respectively. The graph is composed of $l_{\mathrm{max}}=6$ \textit{self} and \textit{cross} attention layers with $h=8$ heads each.  Overall, this results in $33$ linear layers. Our model is implemented in \textit{PyTorch} \cite{paszke2017automatic} \textit{v1.6} with \textit{Python 3.7}.

\textbf{Training procedure:}
We process KITTI's \cite{geiger2012we} odometry training sequences $00$ to $10$, using our key-point selection strategy (cp. Sec.~\ref{section:pillar_layer}) by computing the proposed smoothness function (Eq.~\ref{formula_keypoint_selection}). Ground truth correspondences and unmatched sets are generated using KITTI's odometry ground truth. Ground truth correspondences are either key-point pairs with a nearest neighbor distance smaller than $0.1\,\mathrm{m}$ or invisible matches, i.e. all pairs with distances greater $0.5\,\mathrm{m}$ remain unmatched. We ignore all associations with a distance in range $0.1\,\mathrm{m}$ to $0.5\,\mathrm{m}$ ensuring variances in resulting features. We trained our model for $200$ epochs using Adam \cite{kingma2014adam} with a constant learning rate of $10^{-4}$ and a batch size of 16. For the point cloud registration experiment we chose sequence $00 - 05$ for training and sequence $08 - 10$ for evaluation as stated in \cite{Lu_2019_ICCV} and \cite{bai2020d3feat} with frame differences between source and target frame of $1 - 10$ (frame $\Delta = [1, 10]$). For the LiDAR odometry estimation we used sequence $00 - 06$ for training and $07 - 10$ for evaluation similar to \cite{li2019net} with frame $\Delta = [1, 5]$.

\subsection{Point Cloud Registration}
\textbf{Validation metrics:}
For point cloud registration we adopted the experiments as stated in \cite{Lu_2019_ICCV} where we sampled the test data sequences in 30 frame intervals and select all frames within a 5m radius as registration targets. We calculate the transformation error compared to the ground truth poses provided by the KITTI dataset for each frame pair based on Euclidean distance for translation and $\Theta = 2sin^{-1} (\frac{\norm{R - \bar{R}}_{F}}{\sqrt{8}})$ for rotation, with $\norm{R - \bar{R}}_{F}$ being the Frobenius norm of the chordal distance between estimation and ground truth rotation matrix. Subsequently we determine the mean and max values for each metric as well as the average runtime for registering one frame pair for each approach on the complete test dataset.

\begin{figure*}[!t]
\centering
\includegraphics[width=1\textwidth]{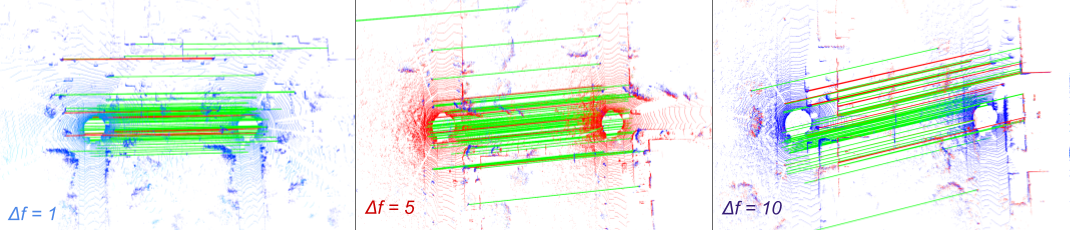}
\caption{\textbf{Qualitative Results} from two point clouds with increasing frame $\Delta$, i.e., increasing difficulty, of $\Delta=1$ (blue - top row), $\Delta=5$ (red - middle row), and $\Delta=10$ (purple - bottom row) frames. The ground truth as well as the model were computed as described in the experiments section. The figure shows samples of the validation sets unseen during training from a different sequence. Green lines highlight correct matches, while red lines highlight incorrect ones.}
\label{fig:qualitative_results}
\end{figure*}

\textbf{Comparison to state-of-the-art methods:}
For pose estimation based on the predicted correspondences by our network, we filter matches below a confidence threshold (e.g. 0.6) and subsequently apply singular value decomposition to determine the pose. Figure \ref{fig:qualitative_results} shows predicted correspondences by the network on an unseen pair of frames for different temporal distances of the target frame. We validate the performance of StickyPillars comparing it to state-of-the-art geometric approaches like ICP~\cite{10.1109/34.121791}, G-ICP~\cite{segal2009generalized}, AA-ICP~\cite{pavlov2018aa}, NDT-P2D~\cite{stoyanov2012fast} and CPD~\cite{myronenko2010point} and the DNN based methods 3DFeat-Net~\cite{yew20183dfeat}, DeepVCP~\cite{Lu_2019_ICCV} and D3Feat~\cite{bai2020d3feat} based on the correspondences predicted by the network. We adopted the results presented in \cite{Lu_2019_ICCV} and extended the table by running the experiments on D3Feat and StickyPillars. For D3Feat we changed the number of key-points to 500 instead of 250 since \cite{bai2020d3feat} reported better performance using this configuration but leaving all other parameters at default. Furthermore according to \cite{Lu_2019_ICCV} the first 500 frame of sequence $08$ involve large errors in the ground truth poses and were therefore neglected during the experiments for point cloud registration. The final results for all methods considered are listed in Table \ref{tab:registration_results}. We are reaching comparable results to all state-of-the-art methods regarding the considered metrics. Moreover we achieve lowest mean angular and second lowest mean translational error for the deep learning based methods without the necessity of an initial pose estimate unlike DeepVCP. In consideration of a mean processing time of 15ms for feature extraction on CPU and an average inference time on a \textit{Nvidia Geforce GTX 1080 Ti} of 101ms for correspondence finding, StickyPillars outperforms all methods regarding the total average runtime for matching one pair of point clouds.

\begin{table}[tbh]
\begin{sc}
\begin{scriptsize}
\begin{center}
\begin{tabular}{l|c c|c c|c}
\toprule
\multirow{2}{*}{Method} & \multicolumn{2}{c|}{Angular Err(°)} & \multicolumn{2}{c|}{Transl. Err(m)} & \multirow{2}{*}{t(s)} \\
 & Mean & Max & Mean & Max &  \\
\midrule
ICP-po2po \cite{10.1109/34.121791}  & 0.139 & 1.176 & 0.089 & 2.017 & 8.17 \\ 
ICP-po2pl \cite{10.1109/34.121791}  & 0.084 & 1.693 & 0.065 & 2.050 & 2.92 \\
GICP \cite{segal2009generalized}    & \textbf{0.067} & \textbf{0.375} & \textbf{0.065} & 2.045 & 6.92 \\
AA-ICP \cite{pavlov2018aa}          & 0.145 & 1.406 & 0.088 & 2.020 & 5.24 \\
NDT-P2D \cite{stoyanov2012fast}     & 0.101 & 4.369 & 0.071 & 2.000 & 8.73 \\
CPD \cite{myronenko2010point}       & 0.461 & 5.076 & 0.804 & 7.301 & 3241 \\
3DFeat-Net \cite{yew20183dfeat}     & 0.199 & 2.428 & 0.116 & 4.972 & 15.02 \\
DeepVCP \cite{Lu_2019_ICCV}         & 0.164 & 1.212 & 0.071 & \textbf{0.482} & 2.3 \\
D3Feat \cite{bai2020d3feat}         & 0.110 & 1.819 & 0.087 & 0.734 & 0.43 \\
\midrule
Ours                                & 0.109 & 1.439 & 0.073 & 1.451 & \textbf{0.12} \\
\bottomrule
\end{tabular}
\end{center}
\end{scriptsize}
\end{sc}
\caption{Point cloud registration results: Our method shows comparable results to state-of-the-art methods and depicting a much lower average runtime.}
\label{tab:registration_results}
\end{table}

\subsection{LiDAR Odometry}
\textbf{Validation metrics:} For validation on LiDAR Odometry, we are using the KITTI Odometry dataset provided with ground truth poses calculating the average translational RMSE $t_{rel}$ (\%) and rotational RMSE $r_{rel}$ (°/100m) on lengths of 100m-800m errors per scene according to \cite{geiger2012we}.

\begin{figure}[tbh]
\center
\includegraphics[width=0.91\columnwidth]{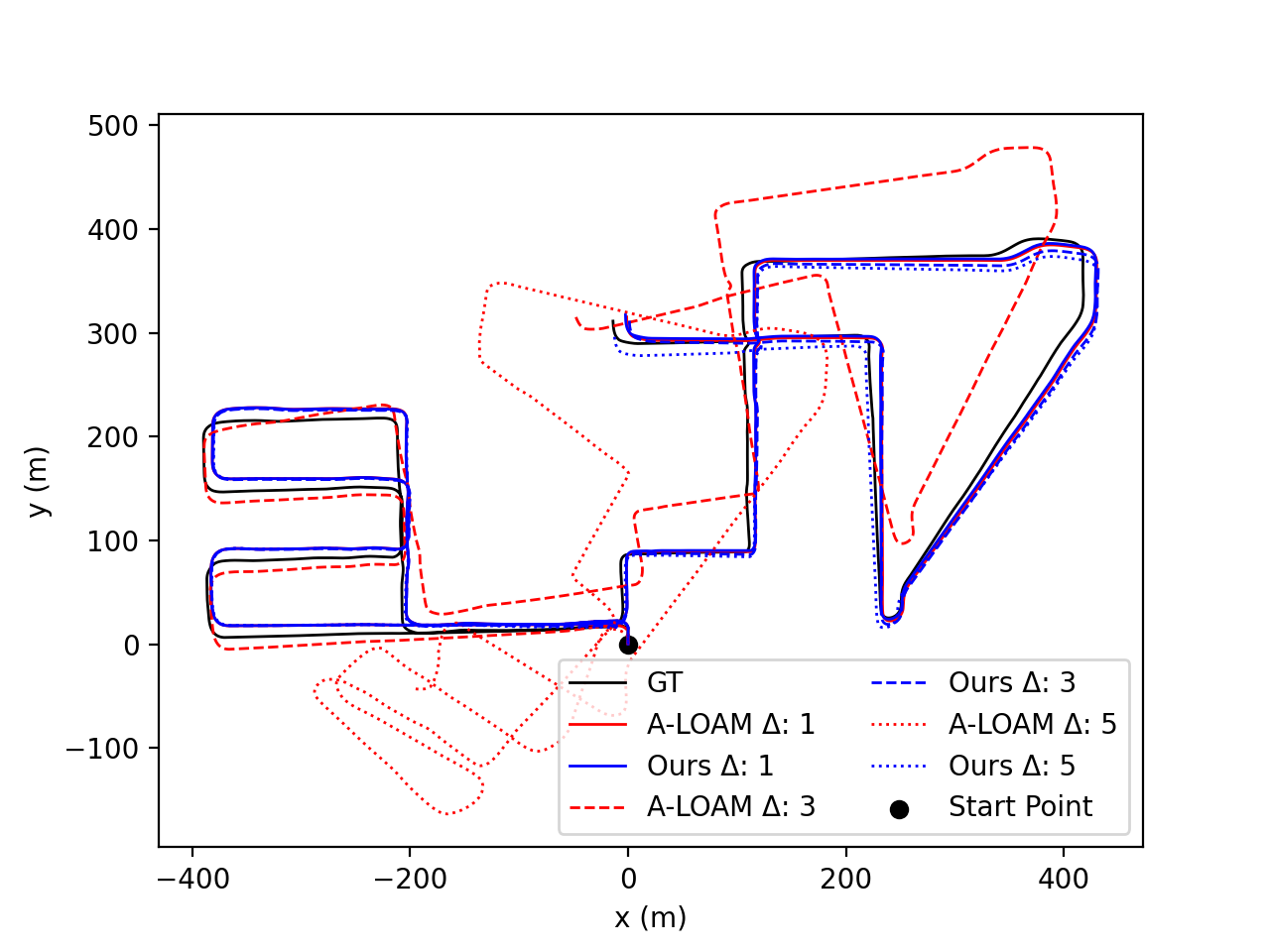}
\caption{\textbf{Trajectory plots} for A-LOAM and A-LOAM + StickyPillars (Ours) for different frame $\Delta$ with ground truth.}
\label{fig:frame_skips}
\end{figure}

\textbf{Comparison to state-of-the-art methods:}
We evaluate the performance of StickyPillars in combination with a subsequent mapping step. For this purpose we utilize the A-LOAM \footnote{https://github.com/HKUST-Aerial-Robotics/A-LOAM} algorithm which is an advanced version of \cite{zhang2017low} and exchange the simple point cloud registration step prior to the mapping with StickyPillars. For our experiments we changed the voxel grid size of the surface features in the mapping step to 1.0m but leaving all other parameters at their default value. In order to achieve real-time capability for LiDAR Odometry, we infer StickyPillars on a \textit{Nvidia Geforce RTX Titan} resulting in a mean run time of 50ms per frame. For all following experiments A-LOAM was processed sequentially, neglecting all kinds of parallel implementations by ROS to ensure a reliable baseline for benchmark comparisons. We compare our results based on the KITTI odometry benchmark to different versions of the ICP algorithm \cite{10.1109/34.121791} \cite{segal2009generalized}, CLS~\cite{velas2016collar} and LOAM~\cite{zhang2017low} which is widely considered as baseline in terms of point cloud based odometry estimation. Furthermore we validate against LO-Net~\cite{li2019net} which is using a similar hybrid approach consisting of a Deep Learning method for point cloud registration and subsequent geometry based mapping. We adopted the values stated in \cite{li2019net} and extended the table with our results as shown in Table \ref{tab:slam_results}. We outperform the considered methods in the majority of sequences regarding $t_{rel}$ and in almost every scene with respect to $r_{rel}$, leading to best results for average translational error on par with LO-Net and lowest average rotational error among all compared approaches.\\
In order to demonstrate the robustness of our method we also compared the performance of the standard A-LOAM implementation to our approach where we replaced the point cloud registration module with StickyPillars in the context of simulating higher speed scenarios and frame drops respectively. This is done by skipping a certain amount of frames of the particular sequence e.g. $\Delta=3$ means providing every third consecutive frame to the algorithm. For evaluation we again use the relative translational and rotational errors as in the previous experiment. For $\Delta=1$, which equals to common processing of a sequence, standard A-LOAM provides some minor improvements to $t_{rel}$ and also $r_{rel}$ in selected scenes where average speed levels are lower, thus leading to large cloud overlap. In such cases the ordinary frame matching algorithm shows good performance. For other sequences like $02$ with a more dynamic environment, the standard implementation fails and the robust transformations provided by StickyPillars help to correct the induced drift leading to much lower average transformation errors. The robustness of our approach in the context of varying vehicle velocities can also be observed by taking a look at the results for higher frame $\Delta$ where, with one exception, our approach outperforms standard A-LOAM in all considered scenes and also depicting comparable average errors to the ones of $\Delta=1$. Furthermore we observed partially better results for larger frame $\Delta$ compared to the smaller ones on certain scenes (e.g. sequence $03$) for our approach which probably is related to a reduction of drift effects caused by close frame to frame matchings. Figure \ref{fig:frame_skips} shows qualitative results of the estimated trajectories for the two methods for different frame $\Delta$ on sequence $08$ which was not seen during the training process of StickyPillars. For $\Delta > 1$ there are large odometry drifts for the standard implementation of A-LOAM whereas the trajectories for the extended version by StickyPillars are almost identical.

\begin{table*}[tbh]
\begin{sc}
\begin{footnotesize}
\begin{center}
\begin{tabular}{c c c c c c c c c c c c c | c c}
\toprule
\multirow{2}{*}{Seq.} & \multicolumn{2}{c}{ICP-po2po~\cite{10.1109/34.121791}} & \multicolumn{2}{c}{ICP-po2pl~\cite{10.1109/34.121791}} & \multicolumn{2}{c}{GICP~\cite{segal2009generalized}} & \multicolumn{2}{c}{CLS~\cite{velas2016collar}} & \multicolumn{2}{c}{LOAM~\cite{zhang2017low}\textsuperscript{1}} & \multicolumn{2}{c|}{LO-Net+Map~\cite{li2019net}} & \multicolumn{2}{c}{Ours} \\
& $t_{rel} $ & $r_{rel}$ & $t_{rel} $ & $r_{rel}$ & $t_{rel} $ & $r_{rel}$ & $t_{rel} $ & $r_{rel}$ & $t_{rel} $ & $r_{rel}$ & $t_{rel} $ & $r_{rel}$ & $t_{rel} $ & $r_{rel}$ \\
\midrule
00\textsuperscript{\dag} &  6.88 &  2.99 &  3.80 &  1.73 &  1.29 &  0.64 &  2.11 &  0.95 &  1.10 (0.78) &  0.53 &  0.78 &  0.42 & \textbf{0.65} & \textbf{0.26} \\
01\textsuperscript{\dag} & 11.21 &  2.58 & 13.53 &  2.58 &  4.39 &  0.91 &  4.22 &  1.05 &  2.79 (1.43) &  0.55 &  \textbf{1.42} &  \textbf{0.40} & 1.82 & 0.45 \\
02\textsuperscript{\dag} &  8.21 &  3.39 &  9.00 &  2.74 &  2.53 &  0.77 &  2.29 &  0.86 &  1.54 (\textbf{0.92}) &  0.55 &  1.01 &  0.45 & 1.00 & \textbf{0.34} \\
03\textsuperscript{\dag} & 11.07 &  5.05 &  2.72 &  1.63 &  1.68 &  1.08 &  1.63 &  1.09 &  1.13 (0.86) &  0.65 &  \textbf{0.73} &  0.59 & 0.91 & \textbf{0.45} \\
04\textsuperscript{\dag} &  6.64 &  4.02 &  2.96 &  2.58 &  3.76 &  1.07 &  1.59 &  0.71 &  1.45 (0.71) &  0.50 &  0.56 &  0.54 & \textbf{0.53} & \textbf{0.17} \\
05\textsuperscript{\dag} &  3.97 &  1.93 &  2.29 &  1.08 &  1.02 &  0.54 &  1.98 &  0.92 &  0.75 (0.57) &  0.38 &  0.62 &  0.35 & \textbf{0.46} & \textbf{0.23} \\
06\textsuperscript{\dag} &  1.95 &  1.59 &  1.77 &  1.00 &  0.92 &  0.46 &  0.92 &  0.46 &  0.72 (0.65) &  0.39 &  \textbf{0.55} &  0.33 & 0.56 & \textbf{0.25} \\
07\textsuperscript{*}    &  5.17 &  3.35 &  1.55 &  1.42 &  0.64 &  0.45 &  1.04 &  0.73 &  0.69 (0.63) &  0.50 &  0.56 &  0.45 & \textbf{0.43} & \textbf{0.24} \\
08\textsuperscript{*}    & 10.04 &  4.93 &  4.42 &  2.14 &  1.58 &  0.75 &  2.14 &  1.05 &  1.18 (1.12) &  0.44 &  1.08 &  0.43 & \textbf{1.02} & \textbf{0.29} \\
09\textsuperscript{*}    &  6.93 &  2.89 &  3.95 &  1.71 &  1.97 &  0.77 &  1.95 &  0.92 &  1.20 (0.77) &  0.48 &  0.77 &  0.38 & \textbf{0.67} & \textbf{0.24} \\
10\textsuperscript{*}    &  8.91 &  4.74 &  6.13 &  2.60 &  1.31 &  0.62 &  3.46 &  1.28 &  1.51 (\textbf{0.79}) &  0.57 &  0.92 &  \textbf{0.41} & 1.00 & \textbf{0.41} \\
\midrule
mean                     &  7.36 &  3.41 &  4.74 &  1.93 &  1.92 &  0.73 &  2.12 &  0.91 &  1.28 (0.84) &  0.51 &  \textbf{0.82} &  0.43 & \textbf{0.82} & \textbf{0.30} \\
\bottomrule
\end{tabular}
\end{center}
\end{footnotesize}
\end{sc}
\begin{footnotesize}
\vspace{-8pt}
\hspace{10pt}\textsuperscript{1}: The results on KITTI dataset outside the brackets are obtained by running the code, and those in the brackets are taken from \cite{zhang2017low}.\\
\hspace*{10pt}\textsuperscript{\dag}: KITTI Odometry dataset sequences used for training\\
\hspace*{10pt}\textsuperscript{*}: KITTI Odometry dataset sequences used for testing
\end{footnotesize}
\caption{LiDAR Odometry results on the KITTI Odometry dataset. We get comparable results regarding $t_{rel}$ and outperfom state-of-the-art methods with respect to $r_{rel}$.}
\label{tab:slam_results}
\end{table*}
\begin{table*}[thb]
\begin{sc}
\begin{footnotesize}
\begin{center}
\begin{tabular}{c | c c c c c c | c c c c c c }
\toprule
\multirow{3}{*}{Seq.} & \multicolumn{6}{c|}{A-LOAM} & \multicolumn{6}{c}{A-LOAM+StickyPillars} \\
 & \multicolumn{2}{c}{$\Delta=1$} & \multicolumn{2}{c}{$\Delta=3$} & \multicolumn{2}{c|}{$\Delta=5$} & \multicolumn{2}{c}{$\Delta=1$} & \multicolumn{2}{c}{$\Delta=3$} & \multicolumn{2}{c}{$\Delta=5$} \\
 & $t_{rel}$ & $r_{rel}$ & $t_{rel}$ & $r_{rel}$ & $t_{rel}$ & $r_{rel}$ & $t_{rel}$ & $r_{rel}$ & $t_{rel}$ & $r_{rel}$ & $t_{rel}$ & $r_{rel}$ \\
\midrule
00\textsuperscript{\dag} &  0.70 & 0.27 & 0.97 & 0.38 & 31.16 & 12.10 & \textbf{0.65} & \textbf{0.26} & \textbf{0.79} & \textbf{0.31} & \textbf{1.29} & \textbf{0.48}\\
01\textsuperscript{\dag} &  1.86 & 0.46 & 4.30 & 0.96 & 96.04 & 10.36 & \textbf{1.82} & \textbf{0.45} & \textbf{2.14} & \textbf{0.59} & \textbf{2.55} & \textbf{0.56}\\
02\textsuperscript{\dag} &  4.58 & 1.43 & 5.29 & 1.61 & 26.74 & 8.72 & \textbf{1.00} & \textbf{0.34} & \textbf{0.91} & \textbf{0.37} & \textbf{1.04} & \textbf{0.42}\\
03\textsuperscript{\dag} &  0.95 & 0.47 & 1.36 & 0.50 & 15.04 & 4.39 & \textbf{0.91} & \textbf{0.45} & \textbf{0.83} & \textbf{0.47} & \textbf{0.77} & \textbf{0.51}\\
04\textsuperscript{\dag} &  0.54 & 0.18 & 89.48 & 0.27 & 102.35 & 0.21 & \textbf{0.53} & \textbf{0.17} & \textbf{0.63} & \textbf{0.25} & \textbf{0.65} & \textbf{0.23}\\
05\textsuperscript{\dag} &  0.47 & 0.24 & 0.53 & 0.24 & 12.47 & 4.21 & \textbf{0.46} & \textbf{0.23} & \textbf{0.52} & \textbf{0.24} & \textbf{0.57} & \textbf{0.27}\\
06\textsuperscript{\dag} &  \textbf{0.55} & \textbf{0.24} & 1.75 & \textbf{0.24} & 42.18 & 12.32 & 0.56 & 0.25 & \textbf{0.57} & 0.26 & \textbf{0.58} & \textbf{0.28}\\
07\textsuperscript{*} & \textbf{0.42} & 0.26 & 0.45 & 0.28 & 11.04 & 3.38 & 0.43 & \textbf{0.24} & \textbf{0.43} & \textbf{0.24} & \textbf{0.49} & \textbf{0.35}\\
08\textsuperscript{*} & \textbf{0.96} & 0.30 & 1.85 & 0.62 & 31.41 & 12.39 & 1.02 & \textbf{0.29} & \textbf{0.92} & \textbf{0.30} & \textbf{1.16} & \textbf{0.38}\\
09\textsuperscript{*} & \textbf{0.66} & \textbf{0.24} & 0.78 & 0.30 & 36.67 & 12.03 & 0.67 & \textbf{0.24} & \textbf{0.72} & \textbf{0.28} & \textbf{0.69} & \textbf{0.31}\\
10\textsuperscript{*} & \textbf{0.87} & \textbf{0.34} & 1.34 & 0.46 & 29.10 & 12.16 & 1.00 & 0.41 & \textbf{0.89} & \textbf{0.38} & \textbf{1.41} & \textbf{0.57}\\
\midrule
mean & 1.14 & 0.40 & 9.83 & 0.53 & 39.47 & 8.39 & \textbf{0.82} & \textbf{0.30} & \textbf{0.85} & \textbf{0.34} & \textbf{1.02} & \textbf{0.40}\\
\bottomrule
\end{tabular}
\end{center}
\end{footnotesize}
\end{sc}
\caption{Extensive experiments demonstrating performance under simulated higher speed / frame drop scenarios with various frame $\Delta$. Our approach shows very high robustness in terms of large environment changes compared to the standard point cloud registration used in A-LOAM.}
\label{tab:slam_results_with_skip}
\end{table*}

\section{Conclusion}
We present a novel model for point-cloud registration in real-time using deep learning. Thereby, we introduce a three stage model composed of a point cloud encoder, an attention-based graph and an optimal transport algorithm. Our model performs local and global feature matching at once using contextual aggregation. Evaluating our method on the KITTI odometry dataset, we observe comparable results to other geometric and DNN based point cloud registration approaches but showing a significantly lower runtime.
Furthermore we demonstrated our capability for robust odometry estimation by adding a subsequent mapping step on the KITTI odometry dataset where we outperformed the state-of-the-art methods regarding rotational error and showing comparable results on the translational error. Finally we proved the robustness of our approach in cases of higher speed scenarios and frame drops respectively, by providing the point clouds with various frame $\Delta$. We showed that even by providing every fifth frame of a sequence StickyPillars is still able to predict accurate transformations thus stabilizing pose estimation when used inside LiDAR odometry and mapping approaches.


\clearpage

{\small
\bibliographystyle{ieee_fullname}
\bibliography{egbib}
}

\end{document}